\documentclass[sigplan,screen]{acmart}
\usepackage{multirow}
\usepackage{colortbl} 
\usepackage{xcolor}   
\usepackage{subcaption}
\AtBeginDocument{%
  }

\setcopyright{acmlicensed}
\copyrightyear{2018}
\acmYear{2018}
\acmDOI{XXXXXXX.XXXXXXX}
\acmConference[Conference acronym 'XX]{Make sure to enter the correct
  conference title from your rights confirmation email}{June 03--05,
  2018}{Woodstock, NY}
\acmISBN{978-1-4503-XXXX-X/2018/06}

\begin{document}

\title{OThink-MR1: Stimulating multimodal generalized reasoning capabilities via dynamic reinforcement learning}







\author{Zhiyuan Liu}
\email{liuzhiyuan1@oppo.com}
\affiliation{%
  \institution{OPPO Research Institute}
  \city{Shenzhen}
  \country{China}
}

\author{Yuting Zhang}
\email{yzhang755@connect.hkust-gz.edu.cn}
\affiliation{\institution{The Hong Kong University of Science and Technology (Guangzhou)}
\city{Guangzhou}
  \country{China}}

\author{Feng Liu}
\email{liufeng4hit@gmail.com}
\affiliation{%
  \institution{OPPO Research Institute}
  \city{Shenzhen}
  \country{China}
}

\author{Changwang Zhang}
\email{zhangchangwang@oppo.com}
\authornote{corresponding authors.}
\affiliation{%
  \institution{OPPO Research Institute}
  \city{Shenzhen}
  \country{China}
}

\author{Ying Sun}
\email{yings@hkust-gz.edu.cn}
\authornotemark[1]
\affiliation{\institution{The Hong Kong University of Science and Technology (Guangzhou)}
\city{Guangzhou}
  \country{China}}
  
\author{Jun Wang}
\authornotemark[1]
\email{junwang.lu@gmail.com}
\affiliation{%
  \institution{OPPO Research Institute}
  \city{Shenzhen}
  \country{China}
}

\renewcommand{\shortauthors}{Zhiyuan Liu et al.}

\begin{abstract}
Multimodal Large Language Models (MLLMs) have gained significant traction for their ability to process diverse input data types and generate coherent, contextually relevant outputs across various applications. While supervised fine-tuning (SFT) has been the predominant approach to enhance MLLM capabilities in task-specific optimization, it often falls short in fostering crucial generalized reasoning abilities. Although reinforcement learning (RL) holds great promise in overcoming these limitations, it encounters two significant challenges: (1) its generalized capacities in multimodal tasks remain largely unexplored, and (2) its training constraints, including the constant Kullback-Leibler divergence or the clamp strategy, often result in suboptimal bottlenecks. To address these challenges, we propose OThink-MR1, an advanced MLLM equipped with profound comprehension and reasoning capabilities across multimodal tasks. Specifically, we introduce Group Relative Policy Optimization with a dynamic Kullback-Leibler strategy (GRPO-D), which markedly enhances reinforcement learning (RL) performance. For Qwen2-VL-2B-Instruct, GRPO-D achieves a relative improvement of more than 5.72\% over SFT and more than 13.59\% over GRPO in same-task evaluation on two adapted datasets. Furthermore, GRPO-D demonstrates remarkable cross-task generalization capabilities, with an average relative improvement of more than 61.63\% over SFT in cross-task evaluation. These results highlight that the MLLM trained with GRPO-D on one multimodal task can be effectively transferred to another task, underscoring the superior generalized reasoning capabilities of our proposed OThink-MR1 model. 
\end{abstract}


\maketitle

\section{Introduction}
Multimodal Large Language Models~\cite{liang2024survey} integrate and process multiple types of input data, such as text and images, to generate coherent and contextually relevant outputs. These models aim to enhance multimodal understandingby leveraging the strengths of different modalities and have been widely used in applications such as Visual Question Answering~\cite{kuang2024natural,lee2024visual} and Image Captioning~\cite{bucciarelli2024personalizing,li2024improving}.

Compared to unimodal large language models (LLMs), MLLMs must process more complex data structures, making their training process significantly more resource-intensive. Furthermore, the multimodal nature of MLLMs positions them as ideal candidates for deployment in complex and dynamic real-world scenarios.   These factors, in turn, place higher demands on the models' cross-task generalization capabilities--the ability to handle different tasks with different data distributions. 




%
\begin{figure}
    \centering
    \includegraphics[width=\linewidth]{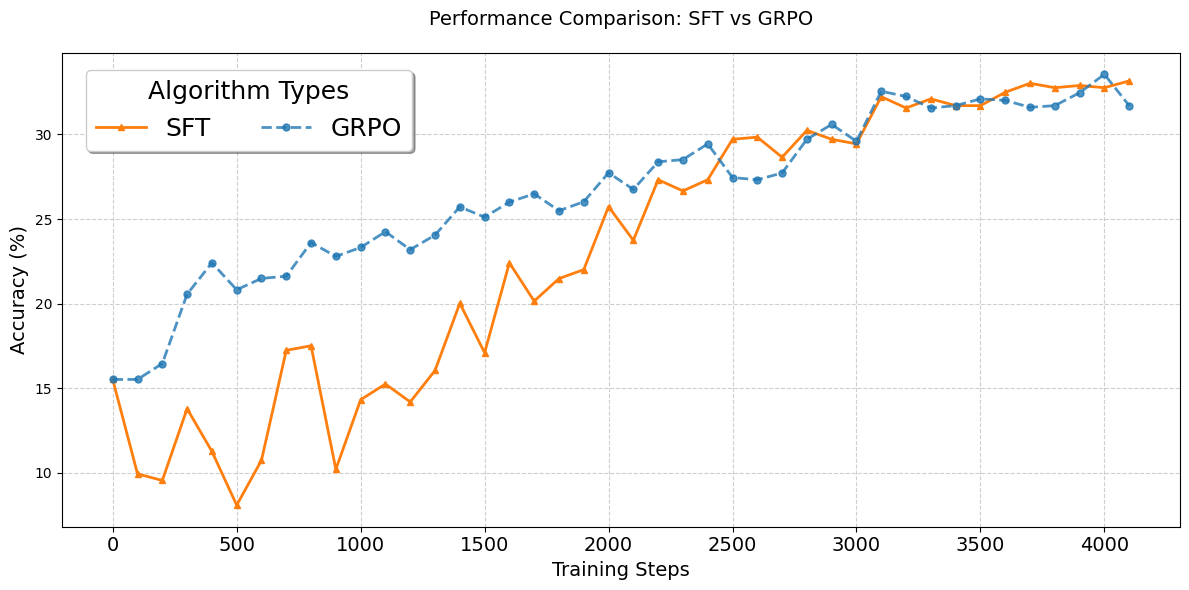}
    \caption{Test accuracy metric curves of SFT and GRPO on geometry reasoning task. }
    \label{fig:performancewtdata}
\end{figure}
Currently, most MLLMs rely on SFT~\cite{lin2024vila,wang2024cogvlm} as the primary post-training approach.  
However, research~\cite{chu2025sft}  has shown  SFT tends to memorize patterns in the training data, performing poorly in out-of-distribution (OOD) scenarios~\cite{chu2025sft}, not to mention cross-task generalization in multimodal settings. This limitation highlights the need for more robust training paradigms that can enhance the generalization capabilities of MLLMs.

Recent advancements in LLMs, such as DeepSeek R1~\cite{guo2025deepseek}, have demonstrated the effectiveness of reinforcement learning with verifiable rewards (RLVR) in improving reasoning abilities. These methods~\cite{team2025kimi,guo2025deepseek} leverage reinforcement fine-tuning  to significantly enhance the reasoning performance of models. However, existing research has primarily focused on unimodal tasks, leaving the potential of RLVR for multimodal tasks and \textbf{its ability to enable cross-task generalization largely unexplored}.

We directly extend reinforcement learning with verifiable rewards to enhance MLLMs  for multimodal tasks. Specifically, we train and validate on the multimodal geometry reasoning task. The test accuracy metric curves of SFT and GRPO (Group Relative Policy Optimization~\cite{guo2025deepseek}, a method based on RLVR)  are shown in Figure~\ref{fig:performancewtdata}. As the number of training steps increases, both SFT and GRPO exhibit performance improvements. However, SFT demonstrates a more significant enhancement and eventually surpasses GRPO when the training steps reach thousands.  \textbf{This can be attributed to GRPO's insufficient exploration, leading to suboptimal solutions}. 

To address these issues, we propose OThink-MR1, an advanced MLLM equipped with  generalized understanding and reasoning across multimodal tasks. 
First, to address \underline{GRPO}'s insufficient exploration capacity, we propose a \underline{D}ynamic KL strategy(GRPO-D) inspired by the $\epsilon$-greedy strategy in classical reinforcement learning. This strategy follows the principle of "early exploration, late exploitation" by dynamically adjusting the KL divergence to balance exploration and exploitation. The validation results demonstrate the dynamic KL strategy effectively enables GRPO-D to surpass SFT in same-task evaluation.

\begin{figure}
    \centering
    \includegraphics[width=0.8\linewidth]{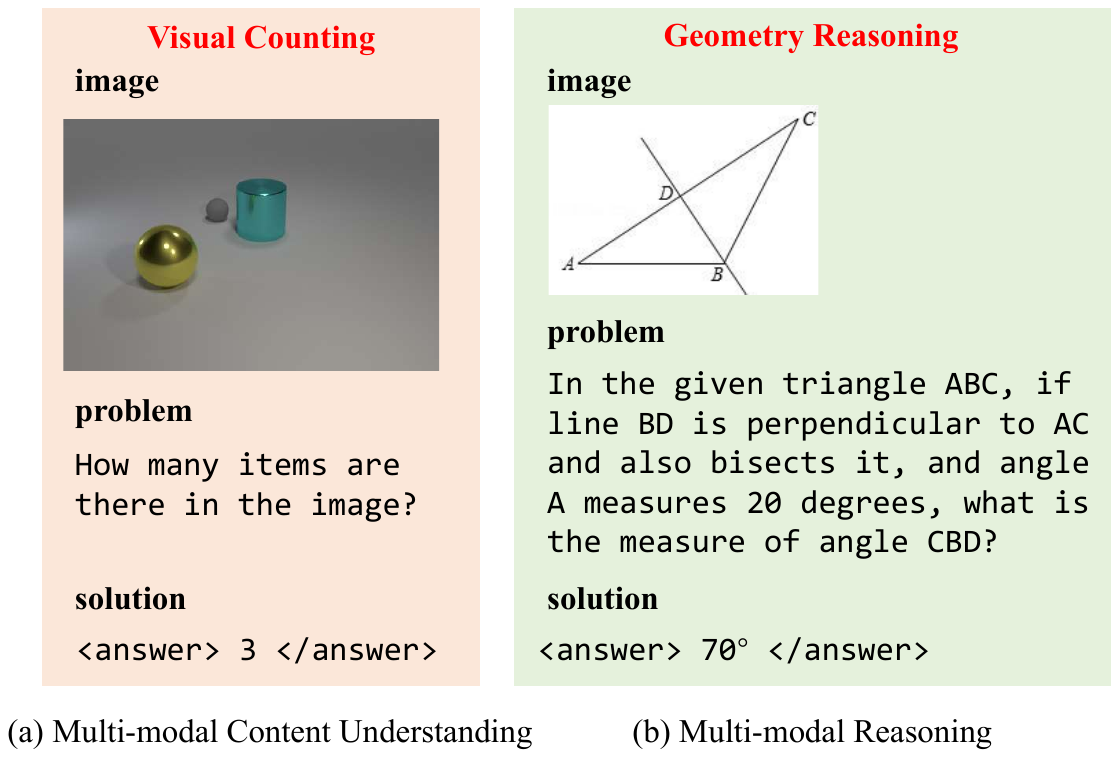}
    \caption{Examples of multimodal content understanding and multimodal reasoning tasks. }
    \label{fig:task_type}
\end{figure}

Additionally, to assess the cross-task generalization ability, we propose cross-task validation, where models are post-trained on one task type and evaluated on another. We adopt the visual counting task (a multimodal content understanding task that extracts basic multimodal information) and the geometry reasoning task (a multimodal reasoning task that requires logical analysis of visual relationships).
The examples of these two tasks are provided in Figure~\ref{fig:task_type}. The cross-task validation results demonstrate the strong generalization capabilities of GRPO-D, highlighting its ability to effectively transfer knowledge across different tasks.

To sum up, our contributions are listed below:
\begin{itemize}
    \item We propose OThink-MR1, a dynamic reinforcement learning framework for fine-tuning MLLMs, which outperforms SFT in the same-task validation. This approach dynamically balances exploration and exploitation, resulting in more effective learning.
   \item We are among the first to demonstrate significant cross-task generalization of dynamic reinforcement learning for MLLMs. This demonstrates that models post-trained with GRPO-D on one multimodal task can be effectively transferred to other multimodal tasks, greatly reducing the need for extensive task-specific data collection and retraining across diverse applications.
  
   \item We conduct extensive experiments on various multimodal tasks, including visual counting, geometry reasoning, cross-task validation and more, to verify our model's generalized reasoning ability.
\end{itemize}

\begin{figure*}
    \centering
    \includegraphics[width=0.8\linewidth]{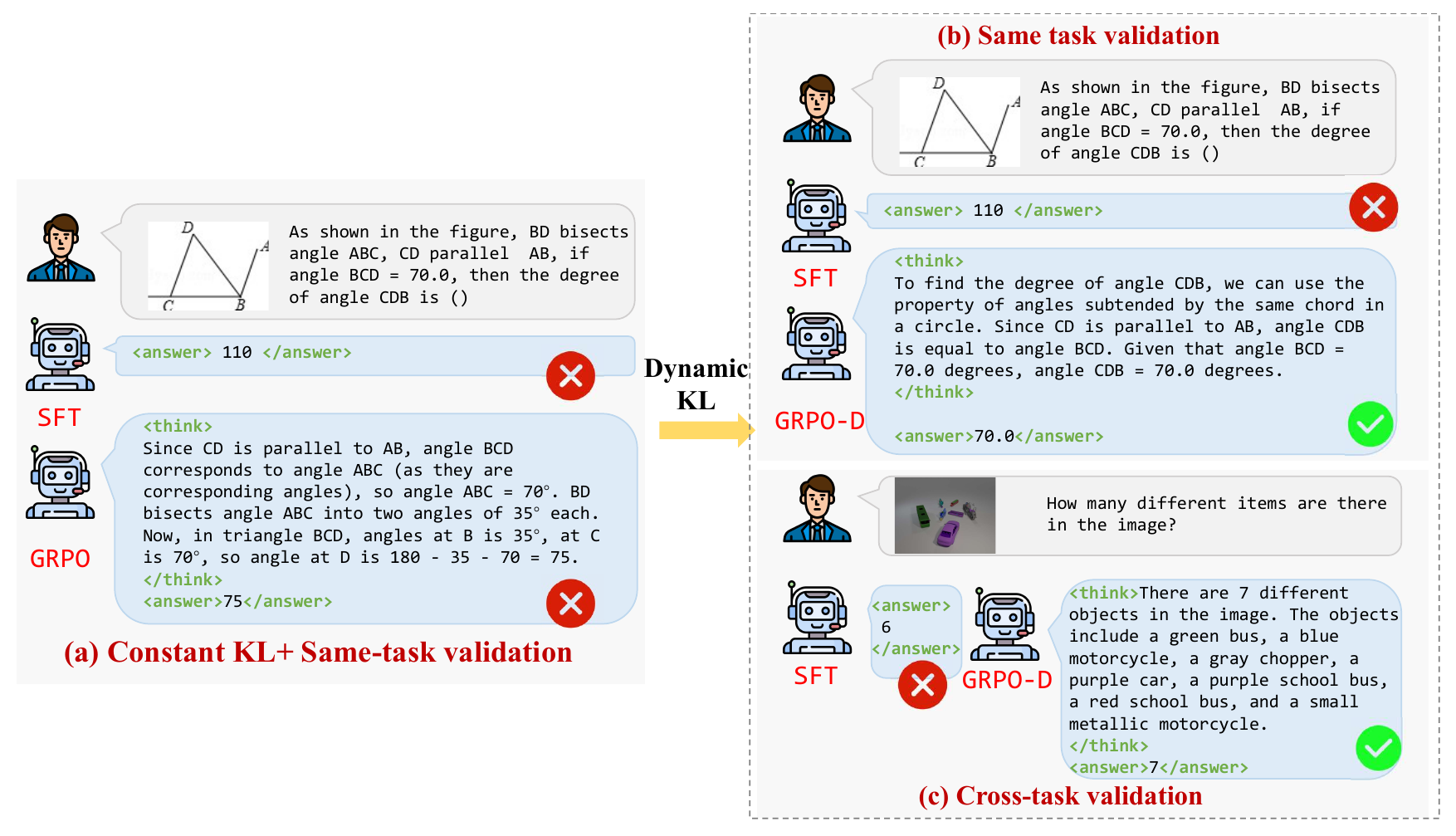}
    \caption{Qualitative illustration of GRPO-D vs SFT in OThink-MR1.}
    \label{fig:enter-label}
\end{figure*}
\section{Related Work}
\textit{Multimodal Large Language Models} such as LLaVA~\cite{liu2023visual} integrate and process multimodal inputs (e.g., text and images) to understand and generate coherent content. Training of MLLMs typically involves pre-training and post-training stages, where post-training is crucial for enhancing model performance. Most works~\cite{lu2022learn,cheng2024domain,zengtimesuite, zhang2023pmc} rely on SFT, while a few~\cite{liu2025visual,chen2025rlvrinvlms} explore RL. However, SFT often struggles with memorization rather than generalization. 
Reinforcement learning (RL), particularly with verifiable rewards (e.g., GRPO-based RLVR in Visual-RFT~\cite{liu2025visual}), excels in few-shot learning scenarios. However, as data scales up, RLVR may be outperformed by SFT due to imbalanced exploration and exploitation, leading to suboptimal solutions. Additionally, prior works, such as R1-V~\cite{chen2025rlvrinvlms}, have evaluated RLVR's out-of-distribution (OOD) performance but are limited to same-task validation, where models are tested only on different data distributions within the same task domain.
To address these issues, we propose a dynamic reinforcement learning algorithm, named GRPO-D. This algorithm adapts an "exploration early, exploitation late" strategy during training, enhancing RLVR's performance and scalability in large-data scenarios where traditional RLVR falls short.
Furthermore, we introduce a novel cross-task validation method. This method underscores GRPO-D's ability to effectively transfer to new multimodal tasks when post-trained on data from another task type, showcasing its ability to foster MLLMs' task-generalized reasoning capability.
\section{Methodology}
In OThink-MR1, we adopt Group Relative Policy Optimization with a Dynamic KL divergence strategy (GRPO-D)  to enhance MLLMs for multimodal tasks. 
The input data for multimodal tasks include images and texts. 
For each multimodal input, GRPO-D samples a group of outputs $\{o_1,o_2,\cdots,o_G\}$ from the old policy $\pi_{\mathrm{old}}$ and  receives corresponding rewards $\{r_1,r_2,\cdots,r_G\}$, which are scored by a verifiable  reward model. In the following, we present the verifiable reward models for visual counting and geometry reasoning tasks.

\subsection{Verifiable Reward for Multimodal Tasks}
In \textbf{Visual counting} and  \textbf{Geometry Reasoning} tasks, we employ a verifiable accuracy reward $R_{acc}$ to assess the numerical or symbolic correctness of the model's output. Specifically, in visual counting, the reward measures the model's output against the ground truth count, whereas in geometry reasoning, it assesses the output against true geometric properties such as angles or lengths.

Also, following DeepSeek-R1-Zero~\cite{guo2025deepseek}, we employ a format reward $R_{format}$ for model optimization. 
This format reward assesses whether the model's output follows the required format, i.e., placing the reasoning process and final answer in the correct tags.
\begin{equation}\label{eq:reward}
    \begin{aligned}
        R = R_{acc} + \alpha R_{format}.
    \end{aligned}
\end{equation}
In prior studies~\cite{liu2025visual,chen2025rlvrinvlms},  the hyper-
parameter $\alpha$ is typically set to 1.  For each response $o_i$, its reward value $r_i$ is calculated  using Eq.~\ref{eq:reward}. The relative quality of responses is then assessed by normalizing these rewards using their mean and standard deviation, as follows:
\begin{equation}
    \begin{aligned}
\hat{A}_{i, t}=\frac{r_i-\operatorname{mean}(\mathbf{r})}{\operatorname{std}(\mathbf{r})}.
    \end{aligned}
\end{equation}

\subsection{Dynamic KL divergence-based Optimization}
After obtaining the relative quality of the output, we need to optimize the policy model $\pi_\theta$. Generally GRPO~\cite{guo2025deepseek} employs a KL  divergence term to constrain the difference between the policy model and the reference model, such as with a default constant weight of 0.04. However, in same-task validation with thousands of samples, GRPO is unable to outperform SFT. We analyze that this may be due to an imbalance between exploration (trying a new policy) and exploitation (maintaining the current effective policy).  A large KL weight restricts exploration early in training, leading to suboptimal performance, while a small weight causes instability due to drastic policy changes later in training.

Inspired by the $\epsilon$-greedy strategy commonly used in Q-learning~\cite{watkins1992q}, a classical reinforcement learning method, we propose a Dynamic KL divergence strategy. Specifically, the  $\epsilon$-greedy strategy adopts the principle of early exploration and late exploitation to ensure robust learning and efficient long-term rewards. Therefore, we introduce a dynamic weight $\beta(s)$ for the KL divergence term, which starts from a small value and gradually increases over the training step $s$. It can be formulated as follows:
\begin{equation}\label{eq:kl_divergence}
    \begin{aligned}
        \beta(s) = 
\begin{cases}
\beta_{\text{mid}} - (\beta_{\text{mid}}-\beta_{min}){} \times \frac{s}{\text{$w$}}, & \text{if } \ s \leq \text{$w$}, \\
\beta_{\text{min}} + (\beta_{\text{max}}-\beta_{min}) \times \frac{s - \text{w}}{\text{$t$ - $w$}}, & \text{if }\ \text{$w$} <  s \leq \text{$t$}, \\
\end{cases}
    \end{aligned}
\end{equation}
where $w$ is the predefined number of exploration steps and $t$ is the total number of training steps. $\beta_{\text{min}}$ and $\beta_{\text{max}}$ are the predefined minimum and maximum values of the weight, with $\beta_{\text{mid}} = (\beta_{\text{min}}+\beta_{\text{max}})/2$. 
Then, we optimize the policy model by maximizing the following objective: 
\small\begin{equation}\label{eq:GRPO_object}
\begin{gathered}
\mathcal{L}_{\mathrm{GRPO}-D}(\theta)=-\frac{1}{G} \sum_{i=1}^G \frac{1}{|o_i|}
\sum_{t=1}^{|o_i|}\\
 [\operatorname{min}(\frac{\pi_\theta(o_{i, t} \mid q, o_{i,<t})}{\pi_{\theta_{\text {old }}}(o_{i, t} \mid q, o_{i,<t})},\operatorname{clip}(\frac{\pi_\theta(o_{i, t} \mid q, o_{i,<t})}{\pi_{\theta_{\text {old }}}(o_{i, t} \mid q, o_{i,<t})}, 1-\epsilon, 1+\epsilon)) \hat{A}_{i, t} \\
-\beta(s) \mathbb{D}_{\mathrm{KL}}[\pi_\theta \mid \pi_{\mathrm{ref}}]], 
\end{gathered}
\end{equation}
\normalsize
where $\epsilon$ is a hyper-parameter, $\pi_{\mathrm{ref}}$ is the reference model before optimization.  
We follow DeepSeek-R1-Zero~\cite{guo2025deepseek} to estimate the KL divergence using an unbiased estimator as follows:
\small\begin{equation}
    \begin{aligned}
        \mathbb{D}_{\mathrm{KL}}[\pi_\theta \| \pi_{\mathrm{ref}}]=\frac{\pi_{\mathrm{ref}}(o_{i, t} \mid q, o_{i,<t})}{\pi_\theta(o_{i, t} \mid q, o_{i,<t})}-\log \frac{\pi_{\mathrm{ref}}(o_{i, t} \mid q, o_{i,<t})}{\pi_\theta(o_{i, t} \mid q, o_{i,<t})}-1. \\
    \end{aligned}
\end{equation}
\normalsize

Overall, in GRPO-D, reducing $\beta(s)$ in the early stage of training allows for greater exploration of the solution space. Later, increasing $\beta(s)$ helps align the model with the reference model. This balance ensures robust learning and efficient long-term rewards.


\section{Experiments}

In this section, we conduct a series of experiments to analyze the impact of various post-training settings on the performance of MLLMs. Specifically, our experiments address the following key research questions:
\begin{itemize}
    \item \textbf{RQ1} 
    How do the reward term and KL divergence term impact the original GRPO in same-task validation?
    \item \textbf{RQ2}  How does the  GRPO-D affect the model's performance in the same-task validation.
    \item \textbf{RQ3} how does SFT or GRPO-D affect the model’s generalization to different tasks?
\end{itemize}

For all experiments, we adopt Qwen2-VL-2/7B-Instruct~\cite{wang2024qwen2} as the backbone model. 
For dataset usage, we follow the settings from R1-V~\cite{chen2025rlvrinvlms}. Specifically, for the geometry reasoning task, we train the model on the 8k GeoQA-Train dataset and validate it on the 753 GeoQA-Test dataset\footnote{https://huggingface.co/datasets/leonardPKU/GEOQA\_R1V\_Train\_8K}~\cite{gao2023g}. 
For the visual counting task, we train the model on the R1 Distilled 37K ClevR dataset\footnote{https://huggingface.co/datasets/leonardPKU/clevr\_cogen\_a\_train} and validate it on the SuperClevR dataset \footnote{https://huggingface.co/leonardPKU/superclevr/tree/main}. Regarding the crucial hyperparameter settings of dynamic KL divergence in Equation~\ref{eq:kl_divergence}, we set 
$w$ and $t$  proportionally, that is $w/t = 0.3$. For the visual counting task, we set  $\beta_{min}=0.04$ and  $\beta_{max}=0.1$. The larger range of $\beta$ values reflects the need for more precise and conservative exploration in simple visual reasoning tasks, where stronger constraints are required to guide the model towards reliable solutions.
For the geometric reasoning task, we set $\beta_{min}=0.0$ and $\beta_{max}=0.02$. The smaller values were selected to allow for a broader range of exploration, which is beneficial for handling the greater complexity and variability in geometric reasoning tasks, where weaker constraints are needed to encourage more diverse solutions.  The other parameters were set following R1-V~\cite{chen2025rlvrinvlms}.

\begin{figure}
    \centering
    \includegraphics[width=\linewidth]{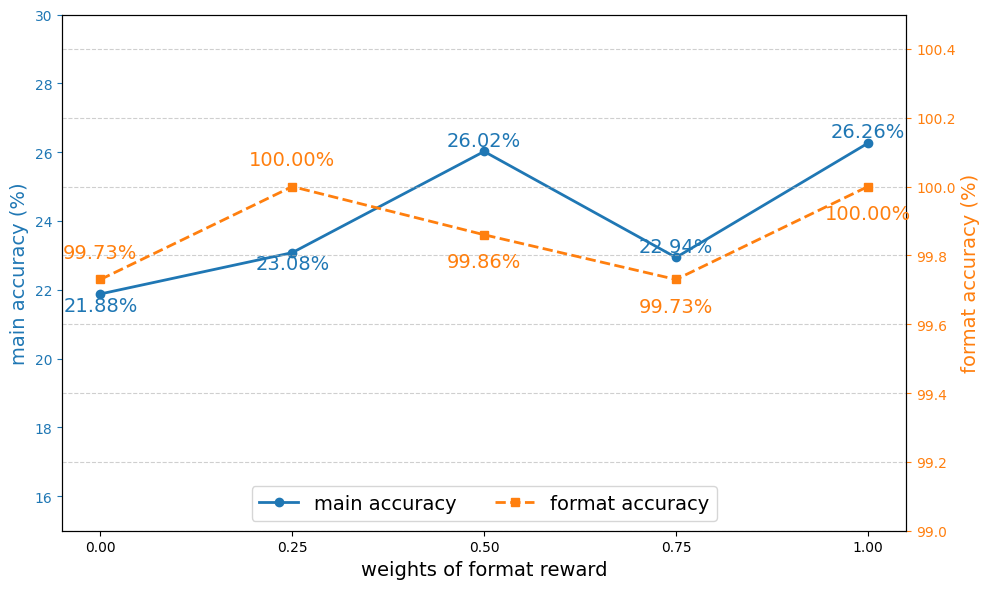}
    \caption{Impact of the weight of format reward.}
    \label{fig:format_reward}
\end{figure}

\subsection{Analysis of GPRO in Multimodal tasks (RQ1)}
    

Eq.~\ref{eq:reward} and Eq.~\ref{eq:GRPO_object} introduce two critical hyper-parameters,
 $\alpha$ and $\beta(s)$ , which control the influence of the format reward and the KL divergence term, respectively, on the model's optimization process. The following analysis examines their individual impacts  for the geometry reasoning task.

\subsubsection{Impact of the weight of format reward} We vary the weight of format reward $\alpha$  in $[0.00,0.25,0.50,0.75,1.00]$. As shown in Figure ~\ref{fig:format_reward}, Qwen2-VL-2B-Instruct+GRPO with format reward (i.e., $\alpha \in [0.25,0.50,0.75,1.00]$) consistently outperforms the version without format reward (i.e., $\alpha=0.0$) on almost all metrics, particularly in terms of accuracy. For instance, when $\alpha=1$, the model achieves a relative improvement of 20.02\%. This improvement is likely due to the combined effect of format reward and accuracy reward, which together guide the model's training more effectively. Even though the format accuracy of the model without format reward was already high (99.70\%), \textbf{explicitly rewarding the correct format can create a more robust and aligned training signal}. This synergy ensures that the model not only generates outputs in the correct format but also does so in a way that enhances its ability to produce accurate answers. 

\begin{figure}[htbp]
    \centering
    \begin{subfigure}[b]{0.4\textwidth}
        \includegraphics[width=\textwidth]{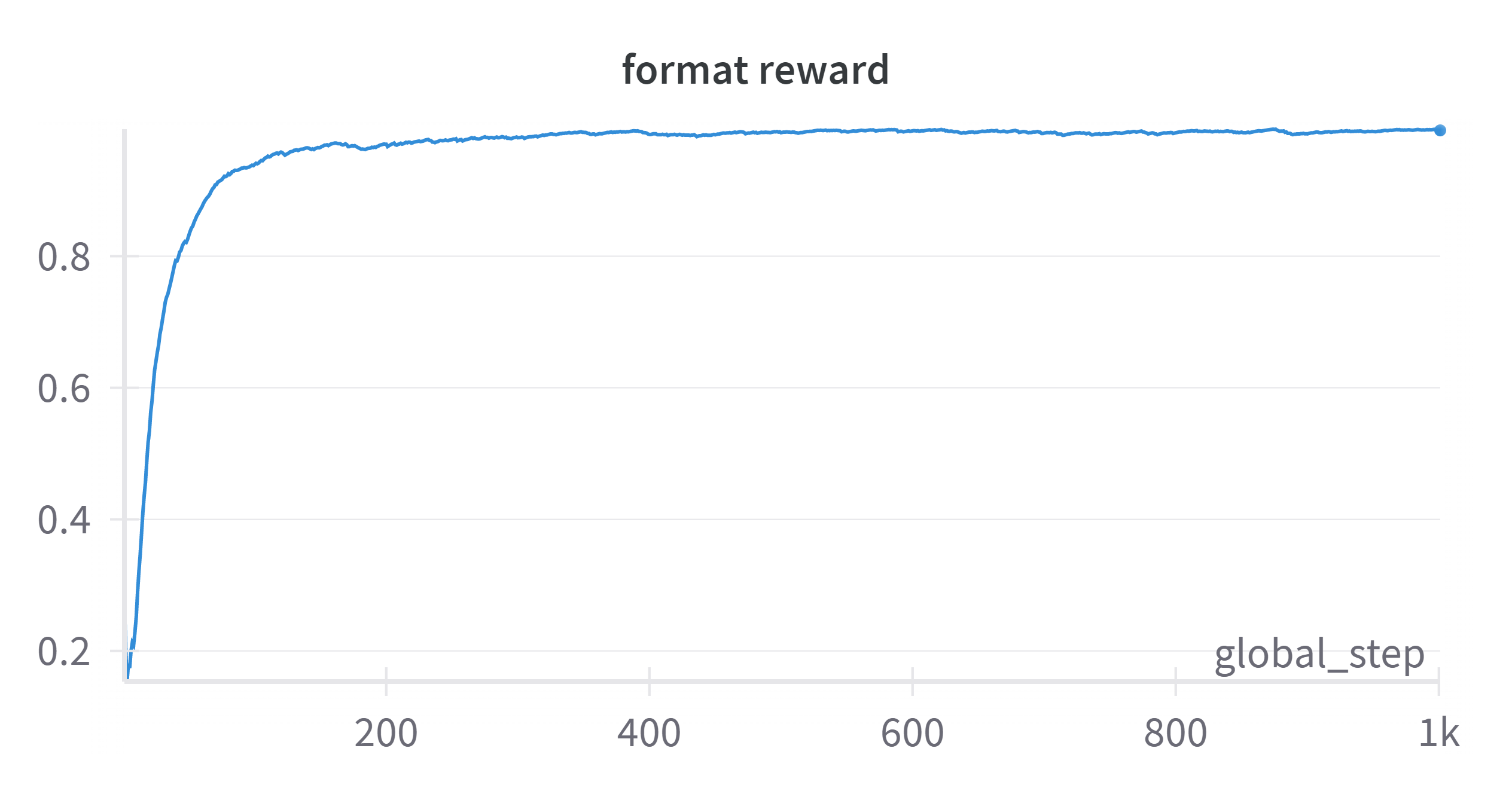}
        \caption{format reward w.r.t training steps.}
    \end{subfigure}
    \begin{subfigure}[b]{0.4\textwidth}
        \includegraphics[width=\textwidth]{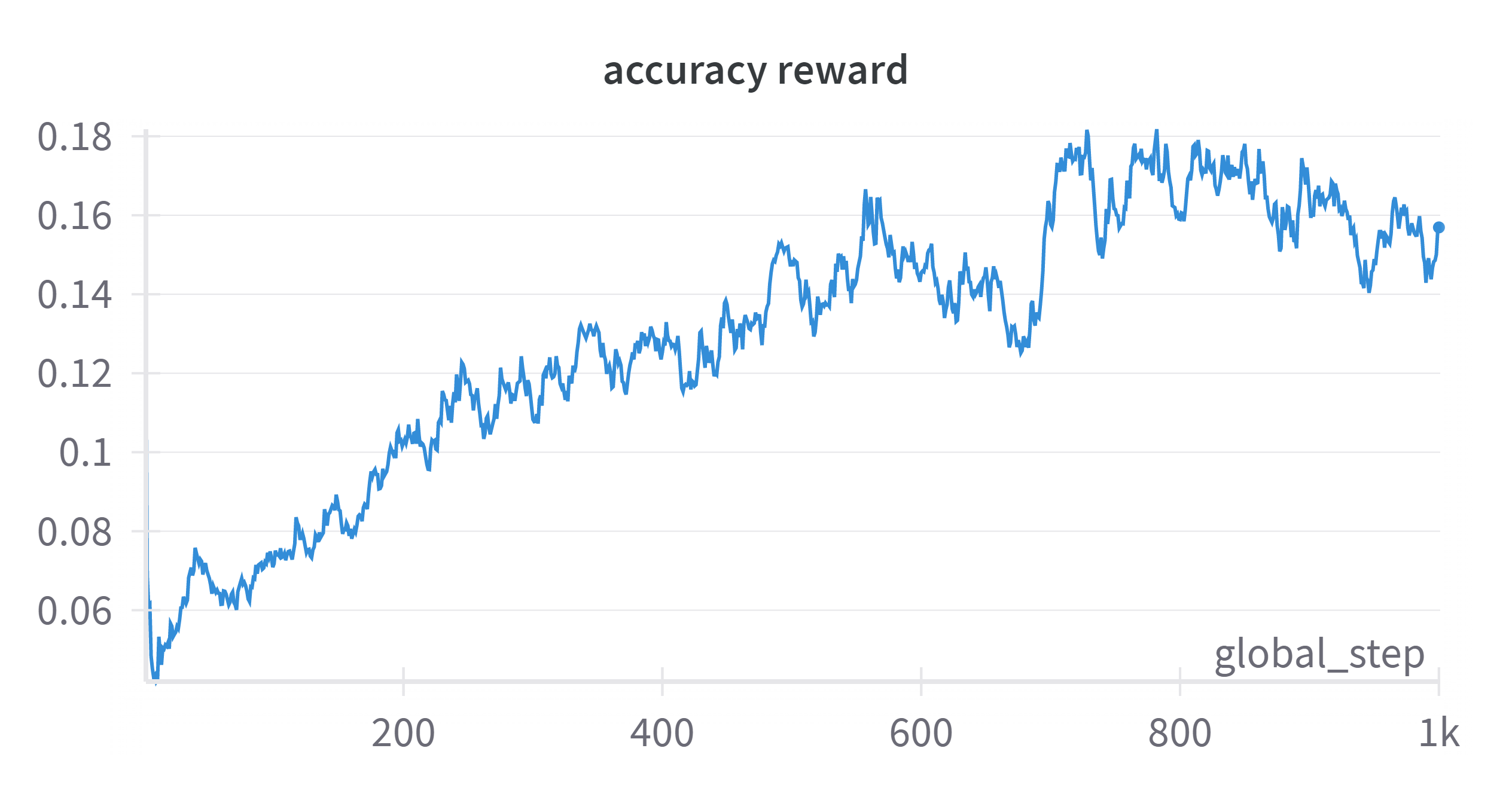}
        \caption{accuracy reward w.r.t training steps.}
    \end{subfigure}
    \caption{Training curves for format reward and accuracy reward.}
    \label{fig:reward_curve}
\end{figure}

Additionally, we present the training curves for accuracy reward and format reward with $\alpha =1$ in Figure~\ref{fig:reward_curve}. The format reward quickly increases from 0.0 to 0.2 within the first 50 steps, while the accuracy reward shows a slow and steady upward trend in the visual reasoning task. This pattern is expected, as \textbf{mastering the reasoning process and generating accurate answers is significantly more challenging than simply aligning the format}.


\subsubsection{Impact of the weight of KL divergence}\label{sec:hy_KL}
\begin{figure}
    \centering
    \includegraphics[width=0.8\linewidth]{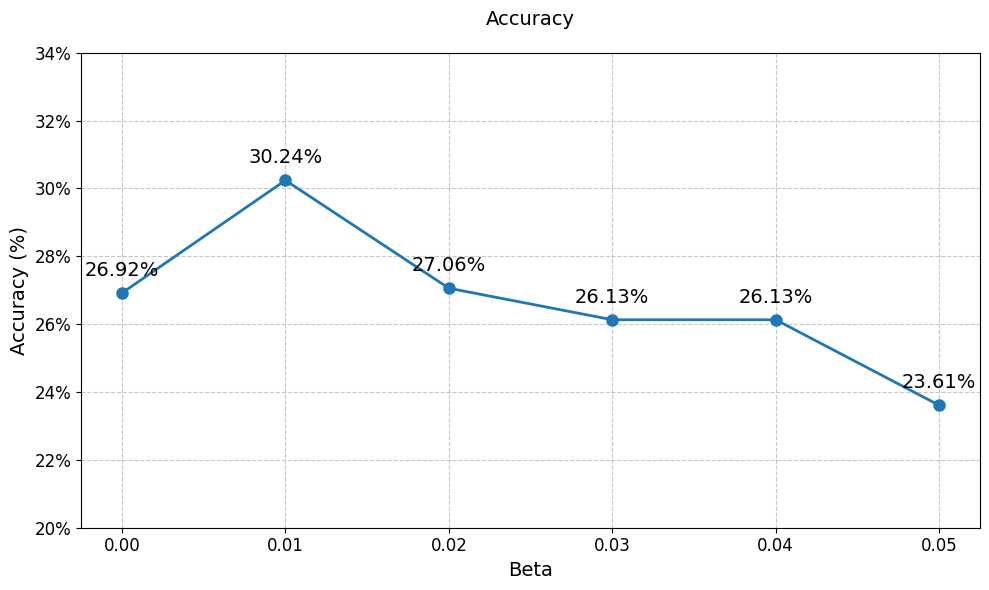}
    \caption{Impact of the weight of KL divergence}
    \label{fig:impact_KL}
\end{figure}

We vary the weight of KL divergence $\beta(s)$  across a range of values $[0.00,0.01,0.02,0.03,0.04,0.05]$ for geometry reasoning task. In each experiment, $\beta(s)$ is held constant throughout training. From the results presented in Figure~\ref{fig:impact_KL}, we can observe that: as the weight of KL divergence increases, model performance first improves and then drops. This occurs because moderate regularization helps in enhancing exploration, but excessive regularization overly constrains the model, leading to underfitting and performance degradation. 

\subsection{Same-Task Evaluation (RQ2)}
However, although adjusting the constant weight of different KL divergence terms can improve the effectiveness of MLLMs in same-task validation, as discussed in subsection~\ref{sec:hy_KL}, we find that GRPO with a constant KL weight (best-performing value 30.24\%) consistently underperforms compared to SFT in same geometry reasoning task validation (32.49\% shown in Table~\ref{tab:overall_performance}).
Thereby, we introduce GRPO-D, which incorporates a dynamic KL divergence strategy to better balance exploration and exploitation in OThink-MR1.
The overall performance on the two adopted datasets for same-task validation is presented in Table~\ref{tab:overall_performance}. 
In the experiments for the visual counting task, we observe that the performance of MLLMs with post-training methods all initially increases and then drops as the training steps increase. This phenomenon may be due to the simplicity of the visual counting task, which makes it prone to overfitting. To address this issue, we adopt a few-shot learning (i.e., 100  samples for Qwen2-VL-2B-Instruct  and 120 samples for Qwen2-VL-7B-Instruct post-training). This strategy helps prevent overfitting by limiting the amount of training data, ensuring that the model generalizes well to unseen data.
Moreover, to ensure fairness, we use the optimal hyperparameters for GRPO:  $\alpha = 1$, a constant $\beta(s)=0.01$ for geometry reasoning task and  $\alpha = 1$, a constant $\beta(s)=0.04$ for visual counting task.  
\begin{table}[]
    \centering
    \caption{Overall performance results for same-task validation, including the visual counting (VC) task and the geometry reasoning (GR) task. The best-performing method is \textbf{bolded}, and the strongest baseline is \underline{underlined}.}
    \begin{tabular}{l|c|c}
    \toprule
    \multirow{2}{*}{Models}&\multicolumn{2}{c}{Accuracy} \\
    \cline{2-3}
     &VC & GR \\

    \midrule
    \midrule
        Qwen2-VL-2B-Instruct &42.50\% & 15.52\% \\
        + GRPO &64.50\% & 30.24\% \\
        + SFT &\underline{68.50\%} &\underline{32.49\%} \\
        \rowcolor{red!20}+ GRPO-D &\textbf{76.50\%} &\textbf{34.35\%} \\  
    \midrule
        \midrule
        Qwen2-VL-7B-Instruct &76.00\% & 33.16\% \\
        + GRPO &\underline{76.50\%} & 42.04\% \\
        + SFT &74.50\% &\underline{43.50\%} \\
        \rowcolor{red!20} + GRPO-D &\textbf{78.00\%} & \textbf{45.49\%} \\
    \bottomrule
    \end{tabular}\label{tab:overall_performance}
\end{table}
From the results, we can observe that:

\begin{itemize}
\item Qwen2-VL-2/7B-Instruct+GRPO underperforms Qwen2-VL-2/7B-Instruct+SFT on most settings. 
This is primarily because, with sufficient training data, SFT can effectively memorize the comprehensive data distribution, leading to better performance in same-task validation.
\item Qwen2-VL-2/7B-Instruct+GRPO-D outperforms Qwen2-VL-2/7B-Instruct+GRPO with constant KL divergence. For instance, for Qwen2-VL-2B-Instruct, the dynamic KL strategy in GRPO-D achieves relative improvements of \textcolor{red}{+18.60\%} and \textcolor{red}{+13.59\%} for visual counting and geometry reasoning tasks, respectively. 
This improvement is attributed to the dynamic KL strategy's ability to balance exploration and exploitation more effectively.
\item Qwen2-VL-2/7B-Instruct+GRPO-D outperforms Qwen2-VL-2/7B-Instruct+SFT. For example, for Qwen2-VL-2B-Instruct, compared to SFT post-training, GRPO-D achieves relative improvements of \textcolor{red}{+11.68\%} and \textcolor{red}{+5.72\%} on the same tasks.
\item Qwen2-VL-2B-Instruct+GRPO-D outperformed Qwen2-VL-7B-Instruct on both tasks. This suggests that post-training algorithm improvements can be more impactful than simply increasing the parameter scale. Specifically, GRPO-D employs dynamic KL divergence to optimize the exploration-exploitation balance, which appears to mitigate some limitations typically associated with smaller model sizes.

\end{itemize}


    

\subsection{Cross-Task Evaluation(RQ3)}

Previous studies~\cite{liu2025visual,chen2025rlvrinvlms}   typically evaluate out-of-distribution performance within the same task. In contrast, our work assesses the generalized reasoning ability across different types of tasks.   \textbf{This requires model have greatly most generalization ability, as it must deal with the totally different tasks and data distributions that it has never encountered before. }
By doing so, we can robustly evaluate whether the model has truly acquired a deep understanding of multimodal knowledge, rather than simply memorizing the post-training data. 

\subsubsection{Cross Task Overall Performance}
We validate the multimodal generalization ability on multimodal content understanding and reasoning tasks,  i.e., visual counting(VC) and geometry reasoning(GR) tasks respectively. The multimodal reasoning tasks are more challenging than multimodal content understanding as they require complex logical deduction beyond direct multimodal interpretation. 
To assess generalization,  we conduct cross-task experiments in both directions as follows: 
\begin{itemize}
    \item \textbf{Reasoning to Understanding Generalization}: post-trained the Qwen2-VL-2B-Instruct on the geometry reasoning training data and validate the trained model on visual counting testing data. 
    \item \textbf{Understanding to Reasoning Generalization}: post-trained the Qwen2-VL-2B-Instruct on the visual counting training data and validate the trained model on geometry reasoning testing data. 
\end{itemize}

The overall cross-task validation results are presented in Figure~\ref{fig:cross_task} and we can observe that: 

\begin{figure}
    \centering
    \includegraphics[width=0.8\linewidth]{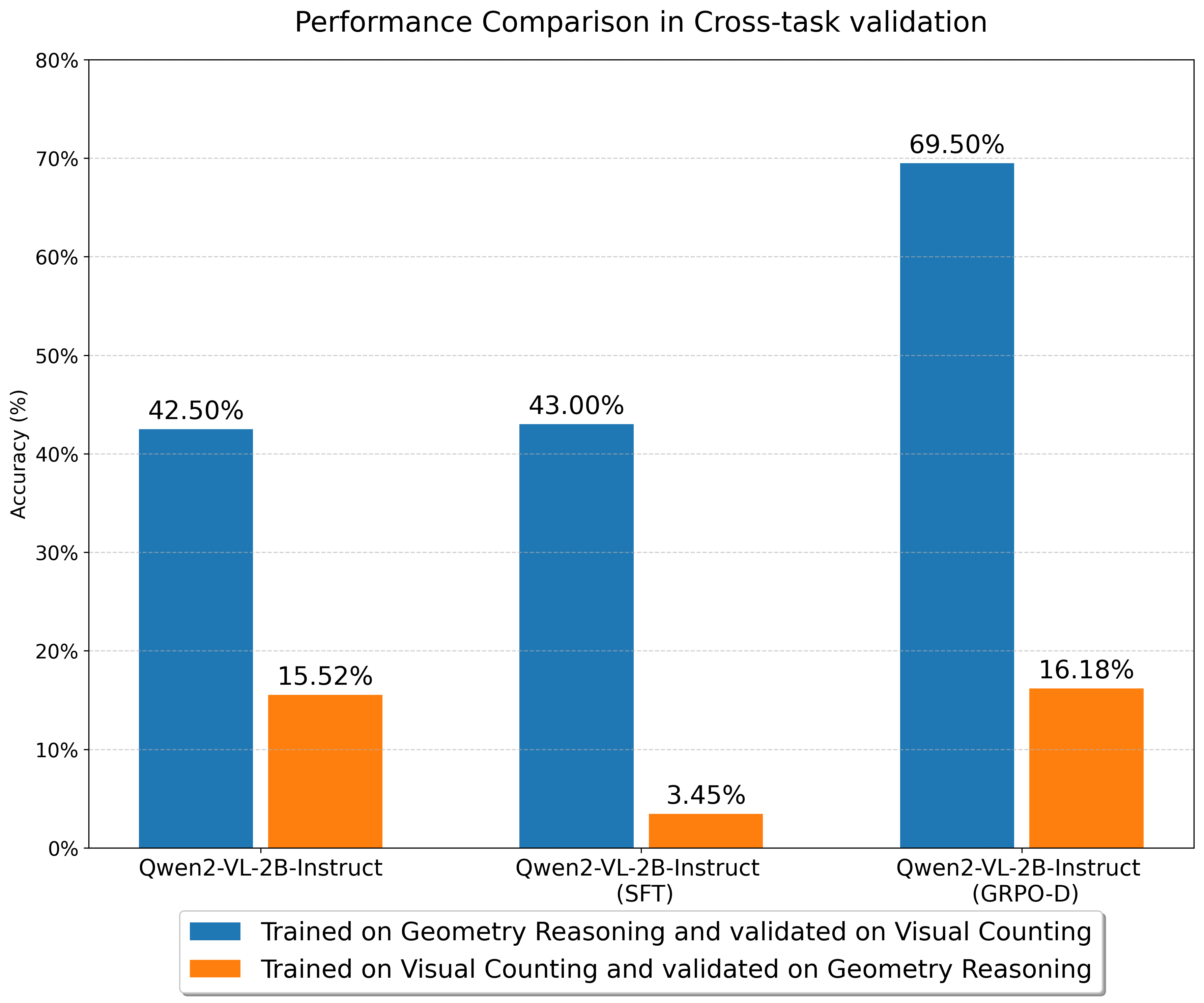}
    \caption{Cross-task validation. }
    \label{fig:cross_task}
\end{figure}

\begin{figure*}
    \centering
    \includegraphics[width=\linewidth]{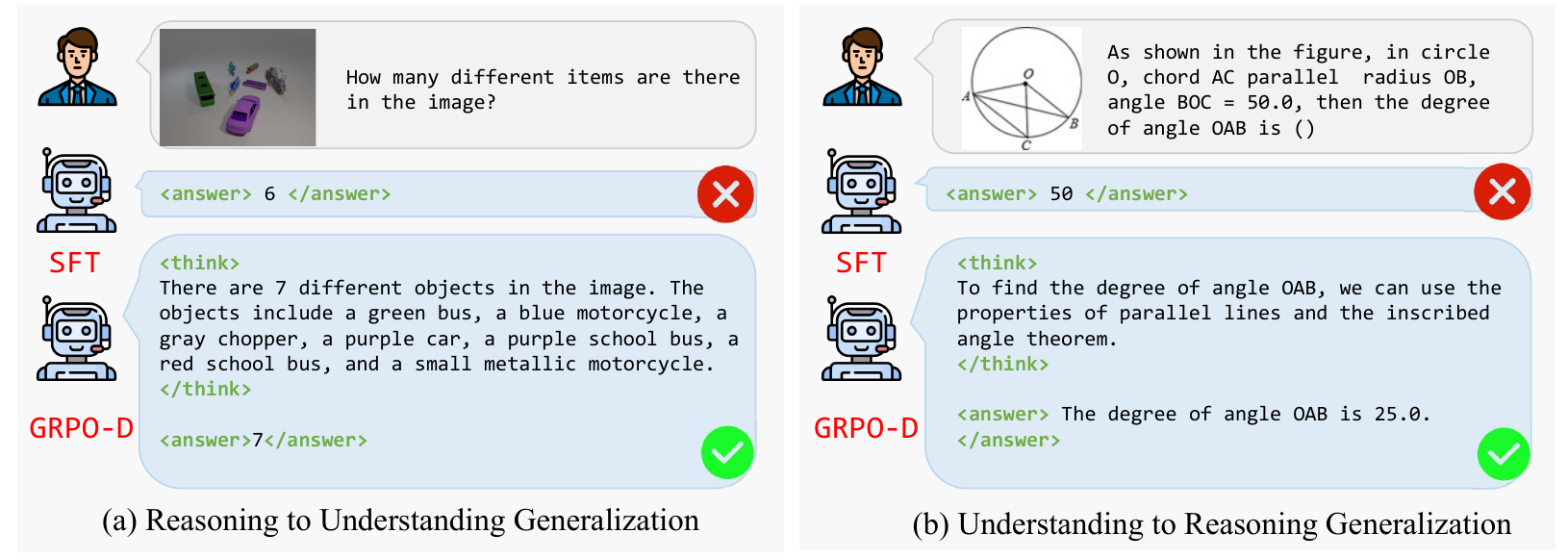}
    \caption{Case studies for cross-task evaluation.}
    \label{fig:case_study}
\end{figure*}
\begin{itemize}
    \item  In the Reasoning to Understanding Generalization experiment, the model with \textbf{SFT} post-training \textcolor{blue}{shows similar performance} to the \textbf{model without post-training}
    In the Understanding to Reasoning Generalization experiment, the model with SFT experiences a relative \textcolor{blue}{-77.77\%} performance drop. These  results indicate that SFT tends to memorize training data rather than learn generalized knowledge, leading to poor cross-task generalization, especially in the more challenging Understanding to Reasoning scenario. 
    \item  In contrast, the model with \textbf{GRPO-D} post-training achieves significant improvements  in both generalization experiments.
    For the Reasoning to Understanding Generalization experiment, it gains a \textcolor{red}{+63.53\%} (vs. without post-training) and \textcolor{red}{+61.63\%} (vs. SFT) 
    relative improvement. For Understanding to Reasoning Generalization experiment, it gains a \textcolor{red}{+4.25\%} (vs. without post-training) and \textcolor{red}{+368.99\%} (vs. SFT) 
    relative improvement.  These results demonstrate that GRPO-D effectively enables the model to acquire task-generalized knowledge in complex multimodal tasks, particularly compared with SFT post-training method. 
    
\end{itemize}
These results are promising, as they demonstrate that \textcolor{red}{\textbf{a model post-trained with GRPO-D on one task can effectively generalize to other multimodal tasks}}. This is particularly evident when the model is post-trained on complex reasoning task. Overall, these results highlight GRPO-D's ability to foster genuine understanding and reasoning of multimodal knowledge, while SFT is often limited to memorization.

\subsubsection{Case Study}
To further evaluate the model's generalized reasoning ability, we perform case studies on cross-task generalization. We select representative samples from both Reasoning to Understanding and Understanding to Reasoning experiments, as shown in Figure~\ref{fig:case_study}. We can observe that after GRPO-D post-training on the geometry reasoning task, the model generalizes effectively to detect and recognize objects in the image, such as identifying the green bus and blue motorcycle. Similarly, in Figure~\ref{fig:case_study}(b), GRPO-D post-training on the visual counting task enables the model to analyze relationships within visual data, such as incorporating the inscribed angle theorem—a key factor contributing to the correct result. These results demonstrate GRPO-D's strong generalized reasoning ability, as evidenced by its adaptable and logical thinking process across diverse tasks.

\section{Conclusion and Discussion}

In this work, we propose OThink-MR1, an advanced MLLM equipped with profound comprehension and reasoning capabilities across multimodal tasks. The key contributions are detailed below:
\begin{itemize}
\item For same-task validation, GRPO initially outperforms SFT in early training stages. However, as training steps increase, SFT matches or surpasses GRPO. This phenomenon occurs because SFT, when exposed to sufficient training data, can memorize the comprehensive distribution of the data. GRPO, on the other hand, struggles with balancing exploration and exploitation, leading to suboptimal performance. To address this, we propose a dynamic reinforcement learning algorithm, named GRPO-D with dynamic KL divergence. Extensive experiments demonstrate that GRPO-D consistently outperforms SFT.
\item When trained and validated on different task types (e.g., multimodal understanding vs. reasoning), GRPO-D significantly outperforms SFT and the base model without post-training. This demonstrates GRPO-D's strong generalized ability to enable MLLMs to achieve deeper understanding and reasoning, highlighting its potential for developing generalized MLLMs applicable to diverse multimodal tasks.
\end{itemize}
Overall, OThink-MR1 enables the development of generalizable multimodal models, representing an advancement in the field of multimodal generalized reasoning. Additionally, several interesting findings have emerged that warrant further exploration: the optimal range of KL weight values appears to be correlated with the complexity of the tasks. Moreover, the generalization ability of GRPO-D appears to be influenced by the reasoning demands placed on the training samples. In the future, we plan to explore these relationships more deeply.

\bibliographystyle{ACM-Reference-Format}
\balance
\bibliography{ref}
\end{document}